\documentclass{article}

\usepackage{microtype}
\usepackage{graphicx}
\usepackage{subfigure}
\usepackage{stfloats}
\usepackage{booktabs} 
\usepackage{pdfpages}
\usepackage{adjustbox}

\usepackage{hyperref}

\usepackage{balance}

\usepackage[accepted]{icml2024}

\usepackage{amsmath}
\usepackage{amssymb}
\usepackage{mathtools}
\usepackage{amsthm}

\usepackage[capitalize,noabbrev]{cleveref}

\theoremstyle{plain}

\theoremstyle{definition}

\theoremstyle{remark}

\newcommand{\makecell}[2][@{}c@{}]{\begin{tabular}{#1}#2\end{tabular}}
\newcommand{\nce}{\mathrm{nce}}

\newcommand{\JinaBTwo}{\href{https://huggingface.co/jinaai/jina-embeddings-v2-base-en}{\texttt{jina-embeddings-v2}}}
\newcommand{\JinaCLIP}
{\href{https://huggingface.co/jinaai/jina-clip-v1}{\texttt{jina-clip-v1}}}

\begin{document}

\setlength{\abovedisplayskip}{8pt}
\setlength{\belowdisplayskip}{8pt}
\setlength{\abovedisplayshortskip}{8pt}
\setlength{\belowdisplayshortskip}{8pt}

\twocolumn[

\icmltitle{\textsc{Jina CLIP}: Your CLIP Model Is Also Your Text Retriever}



\icmlsetsymbol{equal}{*}

\begin{icmlauthorlist}
\icmlauthor{Andreas Koukounas}{equal,comp}
\icmlauthor{Georgios Mastrapas}{equal,comp}
\icmlauthor{Michael G\"unther}{comp}
\icmlauthor{Bo Wang}{comp}
\icmlauthor{Scott Martens}{comp}
\icmlauthor{Isabelle Mohr}{comp}
\icmlauthor{Saba Sturua}{comp}
\icmlauthor{Mohammad Kalim Akram}{comp}
\icmlauthor{Joan Fontanals Martínez}{comp}
\icmlauthor{Saahil Ognawala}{comp}
\icmlauthor{Susana Guzman}{comp}
\icmlauthor{Maximilian Werk}{comp}
\icmlauthor{Nan Wang}{comp}
\icmlauthor{Han Xiao}{comp}
\end{icmlauthorlist}

\icmlaffiliation{comp}{Jina AI GmbH, Ohlauer Str. 43, 10999 Berlin, Germany}
\icmlcorrespondingauthor{Jina AI Research}{research@jina.ai}

\icmlkeywords{Machine Learning, ICML, CLIP, Embeddings, Multimodal, Retrieval}

\vskip 0.3in
]



\printAffiliationsAndNotice{\icmlEqualContribution} 

\begin{abstract}
Contrastive Language-Image Pretraining (CLIP) is widely used to train models to align images and texts in a common embedding space by mapping them to fixed-sized vectors. These models are key to multimodal information retrieval and related tasks. However, CLIP models generally underperform in text-only tasks compared to specialized text models. This creates inefficiencies for information retrieval systems that keep separate embeddings and models for text-only and multimodal tasks. We propose a novel, multi-task contrastive training method to address this issue, which we use to train the \JinaCLIP~model and achieve the state-of-the-art performance on both text-image and text-text retrieval tasks.
\end{abstract}

\section{Introduction}
\label{sec:introduction}
Text-image contrastively trained models, such as CLIP~\cite{clip}, create an aligned representation space for images and texts by leveraging pairs of images and their corresponding captions. Similarly, text-text contrastively trained models, like~\JinaBTwo~\cite{jembeddings2}, construct a representation space for semantically similar texts using pairs of related texts such as question/answer pairs, query/document pairs, or other text pairs with known semantic relationships.

Because image captions are typically very short, CLIP-style models trained with them only support short text context lengths. They struggle to capture the richer information in longer texts, and as a result, perform poorly on text-only tasks. Our empirical study (Table~\ref{tab:results}) demonstrates that OpenAI's CLIP underperforms in all text retrieval tasks. This poses problems for many applications that use larger text inputs, like text-image retrieval, multimodal retrieval augmented generation~\cite{mmrag} and image generation.

In this paper, we present and demonstrate the effectiveness of a novel approach to contrastive training with large-scale image-caption pairs and text pairs.
We jointly optimize for representation alignment of both text-image and text-text pairs, enabling the model to perform well at both kinds of tasks. Due to the lack of available multimodal multi-target datasets (e.g. text-text-image triplets) we use different datasets for each class of task and jointly train for both. 

The resulting model, \JinaCLIP, performs comparably to EVA-CLIP~\cite{evaclip} on the cross-modal CLIP Benchmark\footnote{\url{https://github.com/LAION-AI/CLIP_benchmark}}, while the text encoder by itself performs as well as similar models on MTEB Benchmark tasks~\cite{mteb}.

\begin{figure*}[htbp]
  \centering
  \adjustbox{max width=\textwidth}{\includegraphics{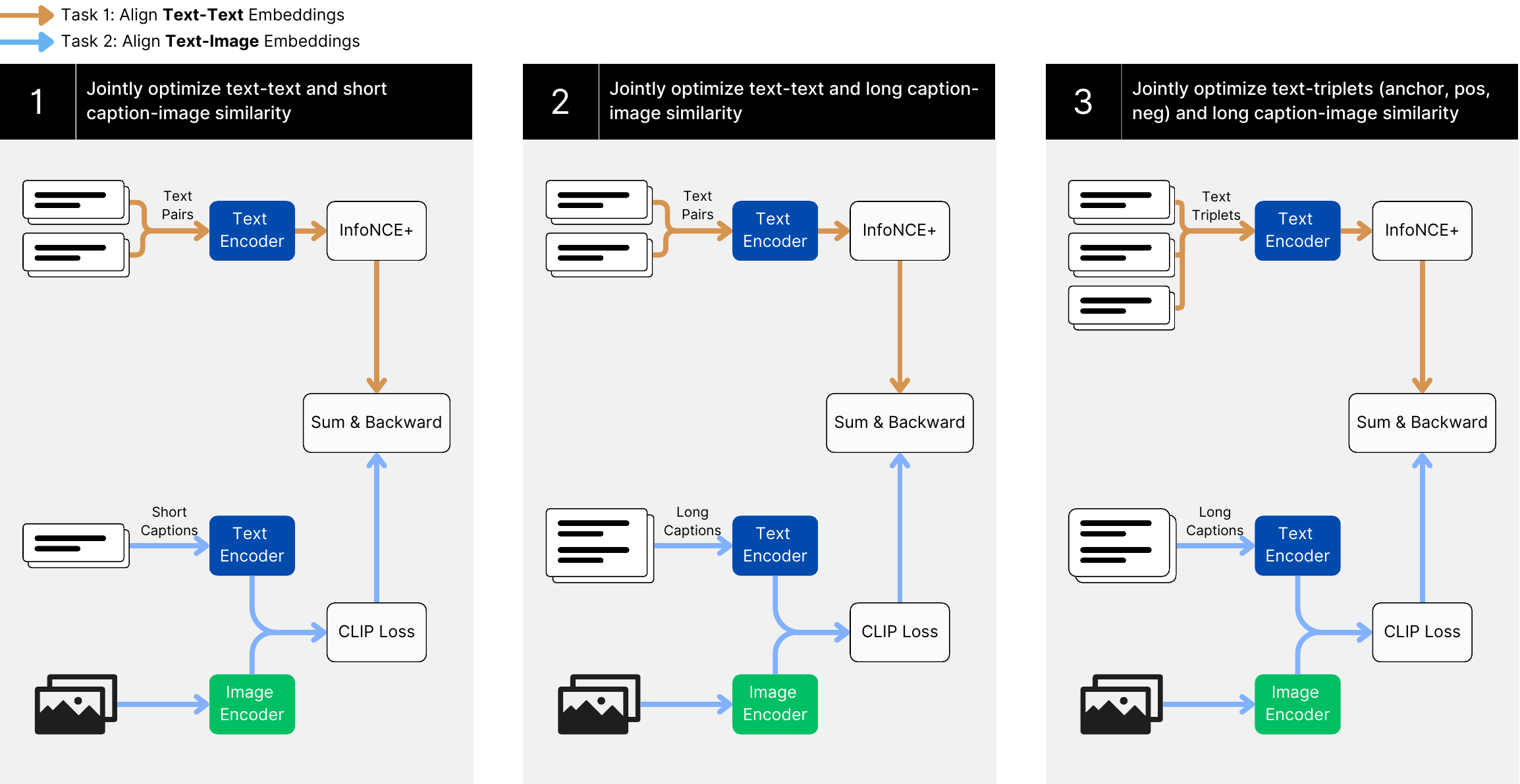}}
  \caption{The training paradigm of \JinaCLIP~, jointly optimizing text-image and text-text matching.}
  \label{figure:training-paradigm}
\end{figure*}

\section{Related Work}
\label{sec:related-work}

\textbf{Contrastive learning for text embeddings} is well-established for training models for text-based information retrieval, semantic textual similarity, text clustering, and re-ranking.
\citet{sbert} propose a dual encoder architecture for pairwise text similarity training.
\citet{dualencoders} demonstrate that the dual-encoder architecture scales efficiently.
\citet{weaklycontrastive} and \citet{jembeddings1} develop multi-stage training methods incorporating \emph{hard negatives}. \citet{jembeddings2multi} bring textual similarity scores directly into the training.
\citet{jembeddings2} and~\citet{bgem3} extend text embedding models' maximum input length to 8,192 tokens.

\textbf{Contrastive text-image pre-training} has become increasingly popular since \citet{clip} proposed the CLIP (Contrastive Language-Image Pre-training) paradigm. Numerous follow-up studies have sought to improve text-image training. 
\citet{lit} introduce \emph{locked image tuning} (LiT), which involves fixing the weights of a trained image encoder and training a text encoder to align with its image representations. 
\citet{3towers} generalize the LiT paradigm to a more flexible \emph{Three Tower} architecture.
\citet{siglip} propose a modified sigmoid loss function for contrastive learning, demonstrating better performance on relatively small batch sizes.
\citet{reproducible} and \citet{evaclip} explore different setups for text-image training, including variations in datasets, model size, and hyperparameters. 
\citet{longclip} empirically determine that the effective context length of CLIP is less than 20 tokens and propose an algorithm to stretch the positional encoding, improving performance on longer texts. 
\citet{evaclip18b} scale up the EVA-CLIP architecture to 18B parameters.

Furthermore, a growing number of large datasets, such as YFCC100M \cite{yfcc100m}, LAION-5B \cite{laion5b}, and curated datasets like ShareGPT4v \cite{sharegpt4v} help to constantly improve the performance of CLIP-like models.

\section{Model Architecture}
\label{sec:model-architecture}

We use the same dual encoder architecture introduced in the original CLIP \cite{clip}.
It comprises a text encoder and an image encoder that generate representations of identical dimensionality.

The text encoder uses the JinaBERT architecture~\cite{jembeddings2}, a BERT variant that integrates AliBi \cite{alibi} to support longer texts.
We pre-train the model using the \emph{Masked Language Modeling} objective from the original BERT model \cite{bert}. Experimental results indicate that this yields superior final performance compared to starting from a text embedding model that has already been fully trained using contrastive learning.

For the image encoder, we use the EVA02 architecture \cite{eva02}. To keep the model size comparable to the text encoder, we select the base variant and initialize our model with the EVA02 pre-trained weights.
Our experiments show that EVA02 significantly outperforms comparable image encoders like DinoV2 \cite{dinov2} and ViT B/16 models from OpenCLIP \cite{openclip}.

\section{Training}
\label{sec:training}

Figure \ref{figure:training-paradigm} illustrates our multi-task, three-stage training approach. This method jointly optimizes the model to perform two tasks: text-image matching and text-text matching.

To achieve text retrieval performance on par with state-of-the-art text embedding models, we employ a multi-stage training paradigm, similar to what \citet{weaklycontrastive} and \citet{jembeddings1} propose. This paradigm introduces a second training stage, analogous to conventional \textit{fine-tuning} stages, that makes use of small high-quality datasets with hard negatives to improve performance on embedding tasks.

We complement the two stages with text-image pairs for a two-task two-stage training pipeline. To address the short effective context length of CLIP-like models, we introduce an intermediate stage for training on long captions. Due to limited availability of public data, we rely on AI-generated long textual descriptions for this stage. The reasoning behind opting for a separate stage is two-fold: a) because of the dataset size difference, integrating this data into the pre-training stage will limit its impact, and b) using the triplet fine-tuning stage for long caption training will limit the effect of hard negatives, as the model is not yet able to effectively process long contexts.

The result is a three-stage training process with two tasks in each stage:
\begin{itemize}
\item \textbf{Stage 1} focuses on learning to align image and text representations while minimizing losses in text-text performance. To this end, we train on large-scale and weakly supervised text-image and text-text pair datasets.
\item \textbf{Stage 2} presents longer, synthetic image captions to the model while continuing to train with text-text pairs.
\item \textbf{Stage 3} uses hard negatives to further improve the text encoder in separating relevant from irrelevant text.
To maintain text-image alignment, we continue training on long image captions.
\end{itemize}

\subsection{Data Preparation}
\label{subsec:data-preparation}

Our text pair corpus $\mathbb{C}^\mathit{text}_\mathit{pairs}$ consists of data from a diverse collection of 40 text-pair datasets, similar to the corpus used in \citet{jembeddings1}.

For text-image training in Stage 1, we use LAION-400M~\cite{laion400m} as our corpus $\mathbb{C}^\mathit{img(s)}_\mathit{pairs}$. LAION-400M contains 400M image-text pairs derived from Common Crawl and is widely used for multimodal training.

In Stages 2 and 3, we use the ShareGPT4V~\cite{sharegpt4v} dataset as our $\mathbb{C}^\mathit{img(l)}_\mathit{pairs}$ corpus.
This dataset contains approximately 100K synthetic captions generated with GPT4v~\cite{gpt4} and an additional 1.1M long captions generated by a large captioning model trained on the original GPT4v generated output. This comes to a total of roughly 1.2M image captions. It would be interesting to investigate the impact of AI-generated data on performance, but that is outside the scope of this work.

Finally, in Stage 3, we use a triplet text corpus $\mathbb{C}^\mathit{text}_\mathit{triplets}$ that includes hard negatives. This corpus combines data from MSMarco \cite{msmarco}, Natural Questions (NQ) \cite{nq}, HotpotQA \cite{hotpotqa} and the Natural Language Inference (NLI) dataset \cite{nli}. Each training batch contains one annotated positive and seven negative items. We select hard negatives using text retrieval models to emphasize relevance in text triplets, except for NLI where negatives are chosen randomly.

\subsection{Loss Functions}
\label{sec:loss-functions}

All three stages employ a joint loss function that combines two InfoNCE loss functions~\cite{infonce}. For the text pairs in stage 1 and stage 2, we use the $\mathcal{L}_{\nce}$ loss function of pairs of text embeddings $(\bf{q},\bf{p}) \sim \mathbf{B}$ within a batch $\mathbf{B} \subset \mathbb{D}^\mathrm{pairs}$. This function evaluates the cosine similarity $cos(\bf{q},\bf{p})$ between a given query $q$ and its corresponding target $p$, relative to the similarity of all other targets in the batch. We sum the loss in both directions to preserve the symmetry of similarity measures:
\begin{flalign}
    & \mathcal{L}_{\nce}(\mathbf{B}) := \mathcal{L}_{\nce}^{\longrightarrow}(\mathbf{B}) + \mathcal{L}_{\nce}^{\longleftarrow}(\mathbf{B}),\text{ with} \nonumber \\
    & \mathcal{L}_{\nce}^{\longrightarrow}(\mathbf{B}) := \mathbb{E}_{(\bf{q},\bf{p})\sim \mathbf{B}}\left[-\ln \frac{e^{cos(\bf{q}, \bf{p})/\tau}}{\sum\limits_{i = 1}^k e^{cos(\bf{q}, \bf{p_i})/ \tau}}\right] \nonumber \\
    & \mathcal{L}_{\nce}^{\longleftarrow}(\mathbf{B}) := \mathbb{E}_{(\bf{q},p)\sim \mathbf{B}}\left[-\ln \frac{e^{cos(\bf{p}, \bf{q}) / \tau}}{\sum\limits_{i = 1}^k e^{cos(\bf{p}, \bf{q_i}) / \tau}}\right]
\end{flalign}
The constant temperature parameter $\tau$ influences how the loss function weighs minor differences in the similarity scores~\cite{understanding}. In accordance with related work~\cite{jembeddings2}, we choose $\tau=0.05$.

Similarly, we apply $\mathcal{L}_{\nce}$ to pairs of caption and image embeddings $(\bf{c},\bf{i}) \sim \mathbf{B}$ in batches $\mathbf{B} \subset \mathbb{D}^\mathrm{img}$ to obtain loss values for text-image matching. For text-image training, $\tau$ is trainable, following the default behaviour in the OpenCLIP framework \cite{openclip}.

For text-text training in stage 3, we use text embeddings from the triplet database $(\bf{q}, \bf{p}, \bf{n_1} ..., \bf{n_7}) \sim \mathbf{B}$ drawn in batches $\mathbf{B} \subset \mathbb{D}^\mathrm{triplets}$.
Recall that these consist of a query $\bf{q}$, a positive match $\bf{p}$, and seven negatives $\bf{n_1} ..., \bf{n_7}$. We employ an extended version of the $\mathcal{L}_{\nce}$ loss, denoted here as $\mathcal{L}_{\nce^+}$, in Equation~\eqref{eq:loss-hard-negatives}. Similar to $\mathcal{L}_{\nce}$, this loss function is bidirectional but incorporates additional negatives when pairing queries with passages:
\begin{flalign}
&\mathcal{L}_{\nce^+}(\mathbf{B}) := \nonumber\\
&\;\;\;\;\;\mathbb{E}_{r\sim \mathbf{B}}\Bigg[-\ln \frac{e^{cos(\bf{q}, \bf{p}) / \tau}}{\sum\limits_{i = 1}^k \Big[ e^{cos(\bf{q}, p_i) / \tau}+ \sum\limits_{j = 1}^{7} e^{cos(\bf{q}, \bf{n_{j,i}}) / \tau}\Big]}\Bigg]\nonumber \\
&\, + \mathbb{E}_{r\sim \mathbf{B}}\Bigg[-\ln \frac{e^{cos(\bf{p}, \bf{q}) / \tau}}{\sum\limits_{i = 1}^k e^{cos(\bf{p}, \bf{q_i}) / \tau}}\Bigg]\nonumber \\
&\text{with}\; r = (\bf{q},\bf{p}, \bf{n_1}, \ldots, \bf{n_{7}}).\label{eq:loss-hard-negatives}
\end{flalign}
\subsection{Training Steps}
\label{sec:training-steps}
In each stage, the text and image encoders are applied to inputs from the corpora described in Section~\ref{subsec:data-preparation} and the training uses the following combinations of loss functions:
\begin{flalign}
&\mathcal{L}_1(\mathbf{B_\mathit{text;s}}, \mathbf{B_\mathit{img;s}}) := \mathcal{L}_{\nce}(\mathbf{B_\mathit{text;s}}) + \mathcal{L}_{\nce}(\mathbf{B_\mathit{img;s}}) \nonumber\\
&\mathcal{L}_2(\mathbf{B_\mathit{text;l}}, \mathbf{B_\mathit{img;l}}) := \mathcal{L}_{\nce}(\mathbf{B_\mathit{text;l}}) + \mathcal{L}_{\nce}(\mathbf{B_\mathit{img;l}}) \nonumber\\
&\mathcal{L}_3(\mathbf{B_\mathit{text3}}, \mathbf{B_\mathit{img;l}}) := \mathcal{L}_{\nce}(\mathbf{B_\mathit{text3}}) + \mathcal{L}_{\nce^+}(\mathbf{B_\mathit{img;l}})
\end{flalign}
For stage 1, $\mathbf{B_\mathit{text;s}}$ is obtained from $\mathbb{C}^\mathit{text}_\mathit{pairs}$ by truncating the text values during tokenization to $77$ tokens as in~\citet{clip}.
This enables us to use very large batches of size $32,768$.
$\mathbf{B_\mathit{img;s}}$ is obtained from $\mathbb{C}^\mathit{img(s)}_\mathit{pairs}$ with the same truncation, albeit most captions in this corpus are short.

For stage 2, $\mathbb{C}^\mathit{text}_\mathit{pairs}$ is used again.
However, text values are truncated to 512 tokens in this case, and as a result a smaller batch size of $8,192$ is used. The text image pairs $\mathbf{B_\mathit{img;l}}$ are selected from $\mathbb{C}^\mathit{img(l)}_\mathit{pairs}$.
During this stage, text-text and text-image retrieval improves by adding synthetic data with longer captions to the training.

The last stage uses text triplets from $\mathbb{C}^\mathit{text}_\mathit{triplets}$ and the text-image batches $\mathbf{B_\mathit{img;l}}$ as in stage 2.
This focused fine-tuning using text triplets and hard negatives brings text-text performance up to competitive levels with specialized text-only models.
Training settings as well as training times for each stage are given in detail in Appendix table \ref{appendix:training-settings}.
\section{Evaluation}
\label{sec:evaluation}
\begin{table*}[t]
    \centering
    \caption{Evaluation results on CLIP Benchmark and MTEB} 
    \label{tab:results}
    \setlength{\tabcolsep}{4.5pt} 
\vskip 0.01in
\small{
\begin{tabular}{ l|rr|rr|r|rr  }
 \toprule
 \textbf{Benchmark} & \multicolumn{2}{c|}{\textbf{CLIP Benchmark}} & \multicolumn{4}{c}{\textbf{MTEB}} \\
 \midrule
 Task Type & \multicolumn{2}{c|}{Zero-Shot Retrieval} & \multicolumn{2}{c}{Retrieval} & STS & Avg MTEB Score \\
 \midrule
 Model - Metric & txt-img r@5 & img-txt r@5 & r@5 & ndcg@10 & spearman & score \\
 \midrule
OpenAI CLIP ViT B/16 & 75.62 & 88.12 & 15.88 & 17.63 & 66.22 & 43.95 \\
EVA-CLIP ViT B/16 & \textbf{82.15} & 90.59 & 22.92 & 26.03 & 69.62 & 47.64 \\
LongCLIP ViT B/16 & 81.72 & \textbf{90.79} & 25.96 & 28.76 & 68.57 & 47.71 \\
\JinaBTwo & - & - & 42.56 & 47.85 & 80.70 & \textbf{60.38} \\
\midrule
\JinaCLIP  stage 1 & 78.05 & 86.95 & 36.29 & 39.52 & 77.96 & 56.51 \\
\JinaCLIP  stage 2 & 81.86 & 90.59 & 36.80 & 40.44 & 78.33 & 57.19 \\
\JinaCLIP & 80.31 & 89.91 & \textbf{43.05} & \textbf{48.33} & \textbf{80.92} & 60.12 \\
 \bottomrule
\end{tabular}

txt-img r@5 : Text to Image Recall@5 [\%]\quad{}
img-txt r@5 : Image to Text Recall@5 [\%]\quad{} 
r@5 : Recall@5 [\%]\quad{} \\
spearman: Spearman Correlation \quad{}
}
\end{table*}
We evaluate our model's performance on text-only tasks, image-only tasks, and cross-modal tasks with both text and images. Table~\ref{tab:results} shows the results of tests comparing \JinaCLIP~to OpenAI CLIP \cite{clip}, EVA-CLIP \cite{evaclip}, and LongCLIP ViT B/16 \cite{longclip} models. Additionally, for text retrieval performance, we include a comparison with \JinaBTwo.
These results demonstrate our model's high performance across all benchmarks.

To evaluate the model's cross-modal performance, we use the CLIP Benchmark which includes zero-shot image-classification and zero-shot cross-modal retrieval tasks. For zero-shot image-text and text-image information retrieval, we evaluate using Flickr8k~\cite{flickr8k}, Flickr30K~\cite{flickr30k} and MSCOCO Captions ~\cite{mscococaptions}, which are all included in CLIP Benchmark. \JinaCLIP ~achieves an average Recall@5 of 85.8\% across all retrieval benchmarks, outperforming OpenAI's CLIP model and performing on par with EVA-CLIP.

To evaluate \JinaCLIP's text encoder, we use the Massive Text Embedding Benchmark (MTEB) \cite{mteb}, which includes eight tasks involving 58 datasets. 
CLIP-like models generally perform poorly on text embedding tasks, particularly information retrieval.
However, \JinaCLIP~ competes closely with top-tier text-only embedding models,  achieving an average score of 60.12\%. This improves on other CLIP models by roughly 15\% overall and 22\% in retrieval tasks.
Detailed results are provided in the appendix.
\section{Conclusion}
\label{sec:conclusion}
We have presented a multi-task, three-stage training method that enables multimodal models to retain high levels of performance on text-only tasks.
The model we produced using this method, \JinaCLIP, exhibits strong performance in cross-modal tasks like text-image retrieval and excels in tasks like semantic textual similarity and text retrieval.
This result confirms that unified multimodal models can replace separate models for different task modalities, with large potential savings for applications.
This model is currently limited to English-language texts due to limited multilingual resources. Future work will focus on extending this work to multilingual contexts.
\balance
\newpage
\bibliography{example_paper}
\bibliographystyle{icml2024}


\appendix
\onecolumn
\section{Appendix}

\begin{table}[h!]
\centering
\caption{Training settings on each stage}
\label{appendix:training-settings}
\vspace{0.1in}
\begin{small}
\begin{tabular}{l|rrr}
\toprule
\textbf{Parameter} & \textbf{Stage 1} & \textbf{Stage 2} & \textbf{Stage 3} \\
\midrule
Image encoder weights init & EVA02 ViT B/16 & Stage 1 & Stage 2 \\
Text encoder weights init. & JinaBERT v2 & Stage 1 & Stage 2  \\
Peak learning rate & 1e-4 & 5e-6 & 1e-6 \\
Image-text pairs batch size & $32,768$ & $8,192$ & $1,024$ \\
Text pairs batch size &  $32,768$ & $8,192$ & $1,024$ \\
Total steps & $60,000$ & $1,500$ & $7,000$ \\
Max sequence length & $77$ & $512$ & $512$ \\
Image-text pairs samples seen & 2B & 12M & 7M \\
Text pairs samples seen & 2B & 12M & 7M \\
Number of GPUs - H100s 80GB & 8 & 8 & 8 \\
Training time including evaluations & 180h & 3h & 4h30m \\
Learning rate schedule & \multicolumn{3}{c}{cosine decay} \\
Optimizer & \multicolumn{3}{c}{AdamW \cite{adamw}} \\
Optimizer hyper-parameters & \multicolumn{3}{c}{\(\beta_1, \beta_2, \epsilon = 0.9, 0.98, 1e-6\)} \\
Weight decay & \multicolumn{3}{c}{0.025} \\
Input resolution & \multicolumn{3}{c}{(224, 224)} \\
Patch size & \multicolumn{3}{c}{(16, 16)} \\
Numerical precision & \multicolumn{3}{c}{AMP} \\
\bottomrule
\end{tabular}
\end{small}
\end{table}

\begin{table*}[htb]
    \centering
    \setlength{\tabcolsep}{4.5pt} 
    \caption{Detailed performance on the CLIP Benchmark}
    \label{appendiX:clip-benchmark}
    \vspace{0.1in}
    \begin{center}
    \begin{small}
    \begin{tabular}{l|cccccc}
        \toprule
        Dataset - Model & \makecell{JinaCLIP} & \makecell{JinaCLIP \\ stage 1} & \makecell{JinaCLIP \\ stage 2} & \makecell{OpenAI CLIP \\ ViT B/16} & \makecell{EVA-CLIP \\ ViT B/16} & \makecell{LongCLIP \\ ViT B/16} \\
        \midrule
        \multicolumn{7}{c}{\textbf{Zero-shot Image Retrieval - Recall@5  [\%]}} \\
        \midrule
        Average & 80.31 & 78.05 & 81.86 & 75.62 & \textbf{82.15} & 81.72 \\
        Flickr30k & 89.02 & 86.88 & 89.80 & 85.60 & \textbf{91.10} & 90.46 \\
        Flickr8k & 85.50 & 84.18 & 87.26 & 82.84 & \textbf{88.50} & 88.40 \\
        MSCOCO  & 66.42 & 63.11 & \textbf{68.54} & 58.42 & 66.85 & 66.31 \\
        \midrule
        \multicolumn{7}{c}{\textbf{Zero-shot Text Retrieval - Recall@5  [\%]}} \\
        \midrule
        Average & 89.91 & 86.95 & 90.59 & 88.12 & 90.59 & \textbf{90.79} \\
        Flickr30k  & 96.50 & 93.80 & 96.10 & 96.20 & 96.60 & \textbf{98.00} \\
        Flickr8k & 94.20 & 90.90 & 94.20 & 91.40 & \textbf{94.60} & 94.00 \\
        MSCOCO  &  79.02 & 76.14 & \textbf{81.38} & 76.76 & 80.58 & 80.38 \\
        \midrule
        \multicolumn{7}{c}{\textbf{Image Classification - Accuracy@1 [\%]}} \\
        \midrule
        Average & 43.28 & 46.74 & 45.39 & 46.16 & \textbf{48.70} & 46.67 \\
        Cars & 68.03 & 76.89 & 69.39 & 64.73 & \textbf{78.56} & 59.17 \\
        Country211 & 13.45 & 15.69 & 13.68 & \textbf{22.85} & 21.34 & 20.28 \\
        Fer2013 & \textbf{49.07} & 38.45 & 47.55 & 46.18 & 51.17 & 47.80 \\
        Fgvc-aircraft & 11.49 & 13.71 & 11.19 & 24.27 &\textbf{ 25.11} & 22.56 \\
        Gtsrb & 38.70 & 41.93 & 39.77 & 43.58 & \textbf{46.33} & 42.93 \\
        Imagenet-a & 29.92 & 33.20 & 30.68 & 49.93 & \textbf{53.89} & 46.84 \\
        Imagenet-o & 33.40 & 32.40 & 34.00 & \textbf{}42.25 & 34.10 & \textbf{42.65} \\
        Imagenet-r & 73.66 & 76.07 & 74.00 & 77.69 & \textbf{82.42} & 76.63 \\
        Imagenet1k & 59.08 & 64.16 & 59.81 & 68.32 & \textbf{74.75} & 66.84 \\
        Imagenet-sketch & 45.04 & 49.33 & 45.90 & 48.25 & \textbf{57.70} & 47.12 \\
        Imagenetv2 & 51.37 & 55.71 & 52.21 & 61.95 & \textbf{66.98} & 60.17 \\
        Mnist & 48.07 & 59.42 & 48.05 & 51.71 & 47.16 & \textbf{71.84} \\\textbf{}
        Objectnet & 45.41 & 51.74 & 45.61 & 55.35 & \textbf{62.29} & 50.79 \\
        Renderedsst2 & 59.14 & \textbf{60.90} & 60.30 & 60.68 & 54.15 & 59.31 \\
        Stl10 & 97.89 & 98.19 & 97.96 & 98.28 & \textbf{99.49} & 98.41 \\
        Sun397 & 65.92 & 68.47 & 65.95 & 64.37 & \textbf{70.62} & 68.73 \\
        Voc2007 & 72.83 & 76.02 & 75.63 & 78.34 & \textbf{80.17 }& 75.35 \\
        Voc2007-multilabel (mean-average-precision [\%]) & 80.62 & 77.94 & 76.80 & 78.91 & \textbf{83.08} & 81.95 \\
        Vtab/caltech101 & 82.68 & \textbf{84.58} & 83.06 & 82.19 & 82.78 & 82.63 \\
        Vtab/cifar10 & 93.49 & 92.68 & 93.83 & 90.78 & \textbf{98.46} & 91.22 \\
        Vtab/cifar100 & 72.08 & 72.62 & 72.67 & 66.94 & \textbf{87.72} & 69.17 \\
        Vtab/clevr-closest-object-distance & 15.61 & \textbf{17.29} & 15.45 & 15.83 & 15.72 & 15.90 \\
        Vtab/clevr-count-all & \textbf{22.35} & 21.53 & 23.49 & 21.09 & 21.27 & 20.71 \\
        Vtab/diabetic-retinopathy & 2.82 & 73.30 & \textbf{73.47} & 3.44 & 14.19 & 10.99 \\
        Vtab/dmlab & 19.53 & \textbf{21.51} & 18.59 & 15.49 & 14.67 & 15.45 \\
        Vtab/dsprites-label-orientation & 2.44 & \textbf{3.33} & 2.86 & 2.34 & 1.94 & 1.12 \\
        Vtab/dsprites-label-x-position & 3.07 & 2.85 & 3.14 & 2.95 & 3.11 & \textbf{3.15} \\
        Vtab/dsprites-label-y-position & 3.17 & \textbf{3.28} & 3.17 & 3.11 & 3.21 & 3.16 \\
        Vtab/dtd & 55.43 & \textbf{56.86} & 55.11 & 44.89 & 52.82 & 45.27 \\
        Vtab/eurosat & 49.52 & 47.00 & 48.35 & 55.93 & \textbf{66.33} & 60.44 \\
        Vtab/flowers & 59.62 & 65.05 & 59.93 & 71.13 & \textbf{75.75} & 69.85 \\
        Vtab/kitti-closest-vehicle-distance & 22.93 & 15.89 & 25.04 & \textbf{26.44} & 22.08 & 34.60 \\
        Vtab/pcam & 55.54 & \textbf{55.79} & 53.30 & 50.72 & 50.95 & 52.55 \\
        Vtab/pets & 80.98 & 86.97 & 80.59 & 89.04 & \textbf{92.10} & 89.21 \\
        Vtab/resisc45 & 55.46 & 57.89 & 54.67 & 58.27 & 60.37 & \textbf{60.63} \\
        Vtab/smallnorb-label-azimuth & \textbf{5.40} & 5.09 & 5.14 & 5.21 & 4.96 & 5.14 \\
        Vtab/smallnorb-label-elevation & 11.31 & 10.98 & 11.24 & \textbf{12.17} & 9.79 & 10.59 \\
        Vtab/svhn & 25.46 & 22.47 & 24.55 & 31.20 & 17.65 & \textbf{27.65} \\
        \bottomrule
    \end{tabular}
\end{small}
\end{center}
\end{table*}

\begin{table*}[htb]
    \centering
    \setlength{\tabcolsep}{4.5pt} 
    \caption{Performance of \JinaCLIP on MTEB Benchmark}
    \vspace{0.1in}
    \label{tab: mteb-averages}
    \small{
    \begin{tabular}{ l|cccccccccc  }
     \toprule
     Model&   CF & CL & PC & RR & RT & STS & SM & Average \\
     \midrule
    OpenAI CLIP ViT B/16  & 60.11 & 35.49 & 71.68 & 46.54 & 17.13 & 66.22 & 29.47 & 43.95 \\
    EVA-CLIP ViT B/16&  60.96 & 37.67 & 74.91 & 47.91 & 25.41 & 69.62 & 28.39 & 47.64 \\
    LongCLIP ViT B/16 &  61.72 & 35.20 & 73.15 & 47.03 & 28.05 & 68.57 & 29.58 & 47.71 \\
    \JinaBTwo & \textbf{73.45} & 41.74 & \textbf{85.38} & \textbf{56.98 }& 47.85 & 80.70 & 31.60 & \textbf{60.38} \\
    \midrule
    \JinaCLIP  stage 1 & 67.54 & \textbf{44.57} & 78.07 & 56.99 & 39.52 & 77.96 & 29.51 & 56.51 \\
    \JinaCLIP  stage 2 & 69.45 & 43.76 & 80.03 & 57.26 & 40.44 & 78.33 & 29.09 & 57.19 \\
    \JinaCLIP &  72.05 & 41.74 & 83.85 & 56.79 & \textbf{48.33} & \textbf{80.92} & \textbf{30.49} & 60.12 \\
    
     \bottomrule
    \end{tabular}

CF: Classification Accuracy [\%] \quad{}
CL: Clustering $\mathcal{V}$ measure [\%]\quad{}
PC: Pair Classification Average Precision [\%]\quad{} \\
RR: Reranking MAP [\%]\quad{}
RT: Retrieval nDCG@10\quad{}
STS: Sentence Similarity Spearman Correlation [\%]\quad{} \\
SM: Summarization Spearman Correlation [\%]\quad{}
}
\end{table*}

\begin{table*}[htb]
    \centering
    \setlength{\tabcolsep}{4.5pt} 
    \caption{Detailed performance on the MTEB Classification tasks}
    \vspace{0.1in}
    \label{tab:mteb-classification}
    \begin{center}
    \begin{small}
    \begin{tabular}{l|ccccccc}
    \toprule
    & \multicolumn{7}{c}{ Accuracy [\%]} \\
    Dataset - Model & \makecell{JinaCLIP} & \makecell{Jina \\ Embeddings-v2} & \makecell{JinaCLIP \\ stage 1} & \makecell{JinaCLIP \\ stage 2} & \makecell{OpenAI CLIP \\ ViT B/16} & \makecell{EVA-CLIP \\ ViT B/16} & \makecell{LongCLIP \\ ViT B/16} \\
    \midrule
    Average Classification                 &              72.05 &              \textbf{73.45} &          67.54 &          69.45 &               60.11 &      60.96&     61.72   \\
    AmazonCounterfactualClassification     &              68.16 &              \textbf{74.73} &          59.85 &          60.78 &               59.58 &      60.92&     60.76  \\
    AmazonPolarityClassification           &              \textbf{96.23} &              88.54 &          93.23 &          95.95 &               63.42 &      63.32&     64.26 \\
    AmazonReviewsClassification            &              44.54 &              \textbf{45.26} &          42.26 &          43.25 &               29.39 &      31.33&     31.65 \\
    Banking77Classification                &              83.94 &              \textbf{84.01} &          82.82 &          83.25 &               73.31 &      74.42&     74.79 \\
    EmotionClassification                  &              47.07 &              \textbf{48.77} &          41.16 &          41.24 &               34.58 &      32.65&     37.11 \\
    ImdbClassification                     &              91.75 &              79.44 &          86.02 &          \textbf{93.50 }&               58.66 &      57.29&     57.53 \\
    MTOPDomainClassification               &              92.67 &              \textbf{95.68} &          89.62 &          90.01 &               87.97 &      92.10&     89.88 \\
    MTOPIntentClassification               &              64.58 &              \textbf{83.15} &          58.74 &          60.44 &               63.36 &      65.76&     65.98 \\
    MassiveIntentClassification            &              69.51 &              \textbf{71.93} &          65.60 &          66.47 &               64.19 &      65.22&     65.80 \\
    MassiveScenarioClassification          &              74.44 &              74.49 &          74.54 &          \textbf{74.82} &               73.18 &      73.14&     74.11 \\
    ToxicConversationsClassification       &              70.47 &              \textbf{73.35} &          60.50 &          66.72 &               63.52 &      63.44&     67.13 \\
    TweetSentimentExtractionClassification &              61.22 &              \textbf{62.06} &          56.15 &          56.97 &   \textbf{}            50.12 &      51.96&     51.70 \\
    \bottomrule\textbf{}
    \end{tabular}
    \end{small}
    \end{center}
\end{table*}
\textbf{}

\begin{table*}[htb]
    \centering
    \setlength{\tabcolsep}{4.5pt} 
    \caption{Detailed performance on the MTEB Clustering tasks}
    \vspace{0.1in}
    \label{tab:mteb-clustering}
    \begin{center}
    \begin{small}
    \begin{tabular}{l|ccccccc}
    \toprule
    & \multicolumn{7}{c}{ $\mathcal{V}$ measure} \\
    Dataset - Model & \makecell{JinaCLIP} & \makecell{Jina \\ Embeddings-v2} & \makecell{JinaCLIP \\ stage 1} & \makecell{JinaCLIP \\ stage 2} & \makecell{OpenAI CLIP \\ ViT B/16} & \makecell{EVA-CLIP \\ ViT B/16} & \makecell{LongCLIP \\ ViT B/16} \\
    \midrule
    Average Clustering         &               41.74 &               41.74 &           \textbf{44.57} &           43.76 &                35.49 &       37.67 &      35.20   \\
    ArxivClusteringP2P         &               44.81 &               45.39 &          \textbf{ 46.26} &           45.32 &                31.86 &       34.03&       32.81   \\
    ArxivClusteringS2S         &               37.81 &               36.68 &           \textbf{39.55} &           39.26 &                27.34 &       26.75&        26.81  \\
    BiorxivClusteringP2P       &               34.74 &               37.05 &           \textbf{38.80} &           36.20 &                31.27 &       31.03&        30.07 \\
    BiorxivClusteringS2S       &               30.78 &               30.16 &           \textbf{34.53} &           34.21 &                27.63 &       27.09&        25.35 \\
    MedrxivClusteringP2P       &               30.82 &               32.41 &           \textbf{33.41} &           31.54 &                29.27 &       29.36&        30.30 \\
    MedrxivClusteringS2S       &               27.64 &               28.09 &           \textbf{31.54} &           31.30 &                27.17 &       26.34&        26.72 \\
    RedditClustering           &               56.21 &               53.05 &           \textbf{59.22} &           59.09 &                42.94 &       49.94&        42.94 \\
    RedditClusteringP2P        &               58.43 &              \textbf{ 60.31} &           58.42 &           57.94 &                52.82 &       58.02&        50.69 \\
    StackExchangeClustering    &               60.35 &               58.52 &          \textbf{ 64.16 }&           63.40 &                52.44 &       57.93&        53.25 \\
    StackExchangeClusteringP2P &               33.46 &               \textbf{34.96} &           33.86 &           33.02 &                30.01 &       32.53&        31.06 \\
    TwentyNewsgroupsClustering &               44.08 &               42.47 &           \textbf{50.50} &           50.12 &                37.61 &       41.33&        37.18 \\
    \bottomrule
    \end{tabular}
    \end{small}
    \end{center}
\end{table*}

\begin{table*}[htb]
    \centering
    \setlength{\tabcolsep}{4.5pt} 
    \caption{Detailed performance on the MTEB Pair-Classification tasks}
    \vspace{0.1in}
    \label{tab:mteb-pair-classification}
    \begin{center}
    \begin{small}
    \begin{tabular}{l|ccccccc}
    \toprule
    & \multicolumn{7}{c}{Average precision based on cosine similarity} \\
    Dataset - Model & \makecell{JinaCLIP} & \makecell{Jina \\ Embeddings-v2} & \makecell{JinaCLIP \\ stage 1} & \makecell{JinaCLIP \\ stage 2} & \makecell{OpenAI CLIP \\ ViT B/16} & \makecell{EVA-CLIP \\ ViT B/16} & \makecell{LongCLIP \\ ViT B/16} \\
    \midrule
    Average Pair Classification        &               83.85 &               \textbf{85.38} &           78.07 &           80.03 &                71.68 &       74.91&    73.15 \\
    SprintDuplicateQuestions &               94.17 &               \textbf{95.30} &           89.42 &           90.32 &                87.33 &       90.20&   89.05 \\
    TwitterSemEval2015       &               71.18 &               \textbf{74.74} &           62.08 &           66.39 &                53.04 &       55.36&   55.21 \\
    TwitterURLCorpus         &              \textbf{ 86.20} &               86.09 &           82.70 &           83.38 &                74.68 &       79.18&   75.19 \\
    \bottomrule
    \end{tabular}
    \end{small}
    \end{center}
\end{table*}

\begin{table*}[htb]
    \centering
    \setlength{\tabcolsep}{4.5pt} 
    \caption{Detailed performance on the MTEB ReRanking tasks}
    \vspace{0.1in}
    \label{tab:mteb-reranking}
    \begin{center}
    \begin{small}
    \begin{tabular}{l|ccccccc}
    \toprule
     & \multicolumn{7}{c}{mAP@10} \\
    Dataset - Model & \makecell{JinaCLIP} & \makecell{Jina \\ Embeddings-v2} & \makecell{JinaCLIP \\ stage 1} & \makecell{JinaCLIP \\ stage 2} & \makecell{OpenAI CLIP \\ ViT B1/6} & \makecell{EVA-CLIP \\ ViT B/16} & \makecell{LongCLIP \\ ViT B/16} \\
    \midrule
    Average Reranking                  & 56.79 & \textbf{56.98} & 56.99 & 57.26 & 46.54 & 47.91 & 47.03 \\
    AskUbuntuDupQuestions     &               61.73 &              \textbf{ 62.25} &           61.26 &           61.65 &                51.23 &       52.22 &    52.57\\
    MindSmallReranking     &               31.21 &               30.54 &           31.42 &           \textbf{31.88} &                26.42 &       28.00 &    26.93 \\
    SciDocsRR                 &               81.76 &               83.10 &           \textbf{83.77} &           83.58 &                71.05 &       70.80 &    70.61\\
    StackOverflowDupQuestions &               \textbf{52.47} &               52.05 &           51.50 &           51.93 &                37.44 &       40.61 &    38.01\\
    \bottomrule
    \end{tabular}
    \end{small}
    \end{center}
\end{table*}

\begin{table*}[htb]
    \centering
    \setlength{\tabcolsep}{4.5pt} 
    \caption{Detailed performance on the MTEB Retrieval tasks}
    \vspace{0.1in}
    \label{tab:mteb-retrieval}
    \begin{center}
    \begin{small}
    \begin{tabular}{l|ccccccc}
    \toprule
    & \multicolumn{7}{c}{nDCG@10} \\
    Dataset - Model & \makecell{JinaCLIP} & \makecell{Jina \\ Embeddings-v2} & \makecell{JinaCLIP \\ stage 1} & \makecell{JinaCLIP \\ stage 2} & \makecell{OpenAI CLIP \\ ViT B/16} & \makecell{EVA-CLIP \\ ViT B/16} & \makecell{LongCLIP \\ ViT B/16} \\
    \midrule
    Average Retrieval        &               \textbf{48.33} &               47.85 &           39.52 &           40.44 &               17.13 &       25.41&   28.05 \\
    ArguAna        &               \textbf{49.36} &               44.18 &           39.53 &           48.26 &                15.51 &       23.49&     32.01\\
    ClimateFEVER   &               \textbf{24.81} &               23.53 &           20.38 &           16.92 &                 3.68 &       19.60&     14.24\\
    CQADupstackRetrieval&          \textbf{40.92} &               39.34 &           35.97 &           39.18 &                10.18 &       16.72&    18.23\\
    DBPedia        &               \textbf{36.64} &               35.05 &           28.41 &           30.33 &                14.94 &       25.42&     27.17\\
    FEVER          &               \textbf{76.28} &               72.33 &           57.50 &           46.72 &                33.45 &       59.26&     63.54\\
    FiQA2018       &               38.27 &               \textbf{41.58} &           36.11 &           38.10 &                 5.78 &        7.33&     11.17\\
    HotpotQA       &               \textbf{61.89} &               61.38 &           40.24 &           43.87 &                 9.30 &       21.54&     33.61\\
    MSMARCO        &               36.91 &               \textbf{40.92} &           25.85 &           27.60 &                 9.36 &       13.76&     17.53\\
    NFCorpus       &               \textbf{33.52} &               32.45 &           31.65 &           32.17 &                16.44 &       21.83&     27.21\\
    NQ             &               58.09 &               \textbf{60.04} &           40.07 &           41.23 &                 5.28 &       10.89&     21.20\\
    QuoraRetrieval &               87.88 &               \textbf{88.20} &           81.55 &           84.32 &                76.63 &       82.32&     78.31\\
    SCIDOCS        &               \textbf{20.24} &               19.86 &           20.06 &           20.20 &                 3.46 &        7.40&     9.24\\
    SciFact        &               67.34 &               66.68 &           \textbf{68.77} &           67.85 &                26.29 &       34.84&     34.77\\
    TRECCOVID      &               \textbf{71.61} &               65.91 &           49.26 &           52.15 &                22.60 &       30.43&     26.42\\
    Touche2020     &               21.15 &               \textbf{26.24} &           17.46 &           17.64 &                 4.10 &        6.35&     6.14\\
    \bottomrule
    \end{tabular}
    \end{small}
    \end{center}
\end{table*}

\begin{table*}[htb]
    \centering
    \setlength{\tabcolsep}{4.5pt} 
    \caption{Detailed performance on MTEB Retrieval tasks - Recall@5}
    \vspace{0.1in}
    \label{MTEB-benchmark-retrieval}
    \begin{center}
    \begin{small}
    \begin{tabular}{l|ccccccc}
    \toprule
    & \multicolumn{7}{c}{Recall@5} \\
    Dataset - Model & \makecell{JinaCLIP} & \makecell{Jina \\ Embeddings-v2} & \makecell{JinaCLIP \\ stage 1} & \makecell{JinaCLIP \\ stage 2} & \makecell{OpenAI CLIP \\ ViT B/16} & \makecell{EVA-CLIP \\ ViT B/16} & \makecell{LongCLIP \\ ViT B/16} \\
    \midrule
    Average - R@5      &               \textbf{43.05 }&               42.56 &           36.29 &           36.80 &                15.88 &       22.92 &       25.96 \\

    ArguAna        &               \textbf{62.37} &               53.62 &           48.01 &           59.74 &                18.77 &       27.60 &       37.98 \\
    CQADupstackRetrieval&          \textbf{44.80} &               43.24 &           40.23 &           43.19 &                11.47 &       18.61&        20.26\\
    ClimateFEVER   &               \textbf{23.73} &               22.26 &           19.80 &           16.33 &                 3.38 &       18.57 &       13.33 \\
    DBPedia        &               \textbf{17.82} &               16.61 &           15.37 &           15.71 &                 6.78 &       11.42 &       12.62 \\
    FEVER          &               \textbf{85.93} &               81.67 &           69.75 &           57.61 &                40.62 &       68.57 &       74.02 \\
    FiQA2018       &               38.18 &               \textbf{39.36} &           34.80 &           36.36 &                 5.83 &        7.69 &       11.41 \\
    HotpotQA       &               \textbf{58.95} &               58.55 &           38.18 &           41.96 &                 8.99 &       20.71 &       31.61 \\
    MSMARCO        &               46.16 &               \textbf{49.73} &           32.57 &           34.04 &                11.73 &       16.85 &       21.59 \\
    NFCorpus       &               \textbf{13.04} &               12.41 &           12.67 &           12.93 &                 5.98 &        7.69 &        9.21 \\
    NQ             &               67.36 &               \textbf{70.37} &           48.89 &           50.01 &                 6.31 &       12.69 &       25.68 \\
    QuoraRetrieval &               91.33 &               \textbf{91.69} &           85.21 &           88.06 &                80.54 &       86.21 &       82.31 \\
    SCIDOCS        &               14.85 &               14.64 &           \textbf{14.86} &           14.73 &                 2.57 &        5.16 &        6.51 \\
    SciFact        &               72.11 &               73.27 &           \textbf{74.94} &           73.79 &                33.08 &       38.89 &       40.34 \\
    TRECCOVID      &               \textbf{1.04} &               1.01 &            0.74 &            0.78 &                 0.32 &        0.47 &        0.46 \\
    Touche2020     &               8.04 &               \textbf{9.99} &            8.39 &            6.79 &                 1.79 &        2.67 &        2.10 \\
    \bottomrule
    \end{tabular}
    \end{small}
    \end{center}
\end{table*}

\begin{table*}[htb]
    \centering
    \setlength{\tabcolsep}{4.5pt} 
    \caption{Detailed performance on the MTEB STS tasks}
    \vspace{0.1in}
    \label{tab:mteb-sts}
    \begin{center}
    \begin{small}
    \begin{tabular}{l|ccccccc}
    \toprule
    & \multicolumn{7}{c}{Spearman correlation based on cosine similarity} \\
    Dataset - Model & \makecell{JinaCLIP} & \makecell{Jina \\ Embeddings-v2} & \makecell{JinaCLIP \\ stage 1} & \makecell{JinaCLIP \\ stage 2} & \makecell{OpenAI CLIP \\ ViT B/16} & \makecell{EVA-CLIP \\ ViT B/16} & \makecell{LongCLIP \\ ViT B/16} \\
    \midrule
    Average STS      &               \textbf{80.92} &               80.70 &           77.96 &           78.33 &                66.22 &       69.62&     68.57 \\
    BIOSSES      &              \textbf{ 83.75} &               81.23 &           83.32 &           83.74 &                67.78 &       71.18&      70.44 \\
    SICK-R       &               78.95 &               \textbf{79.65} &           76.76 &           76.77 &                69.08 &       73.72&      72.59 \\
    STS12        &               73.52 &               \textbf{74.27} &           69.52 &           70.97 &                72.07 &       70.19&      72.63 \\
    STS13        &               83.24 &               \textbf{84.18} &           78.03 &           78.15 &                64.44 &       63.02&      66.25 \\
    STS14        &               78.68 &               \textbf{78.81} &           72.44 &           73.20 &                55.71 &       59.98&      58.66 \\
    STS15        &               87.46 &               \textbf{87.55} &           84.39 &           84.51 &                65.37 &       73.12&      68.81 \\
    STS16        &               83.77 &               \textbf{85.35} &           78.70 &           79.27 &                72.44 &       74.74&      72.43 \\
    STS17        &               \textbf{89.77} &               88.88 &           88.44 &           88.10 &                77.23 &       81.90&      79.72 \\
    STS22        &               \textbf{65.15} &               62.20 &           66.45 &           66.64 &                53.63 &       59.33&      55.60 \\
    STSBenchmark &               \textbf{84.93} &               84.84 &           81.57 &           81.96 &                64.40 &       69.01&      68.55 \\
    \bottomrule
    \end{tabular}
    \end{small}
    \end{center}
\end{table*}

\begin{table*}[htb]
    \centering
    \setlength{\tabcolsep}{4.5pt} 
    \caption{Detailed performance on the MTEB Summarization tasks}
    \vspace{0.1in}
    \label{tab:mteb-summarization}
    \begin{center}
    \begin{small}
    \begin{tabular}{l|ccccccc}
    \toprule
    & \multicolumn{7}{c}{Spearman correlation based on cosine similarity} \\
    Dataset - Model & \makecell{JinaCLIP} & \makecell{Jina \\ Embeddings-v2} & \makecell{JinaCLIP \\ stage 1} & \makecell{JinaCLIP \\ stage 2} & \makecell{OpenAI CLIP \\ ViT B/16} & \makecell{EVA-CLIP \\ ViT B/16} & \makecell{LongCLIP \\ ViT B/16} \\
    \midrule
    SummEval &               30.49 &               \textbf{31.60} &           29.51 &           29.09 &                29.47 &       28.39&  29.58 \\
    \bottomrule
    \end{tabular}
    \end{small}
    \end{center}
\end{table*}

\end{document}